# Group Event Detection with a Varying Number of Group Members for Video Surveillance

Weiyao Lin*, Ming-Ting Sun*, Radha Poovendran*, and Zhengyou Zhang**
*Department of Electrical Engineering, University of Washington, Seattle, WA, USA
** Microsoft Research, Microsoft Corp., Redmond, WA, USA

## Abstract

This paper presents a novel approach for automatic recognition of group activities for video surveillance applications. We propose to use a group representative to handle the recognition with a varying number of group members, and use an Asynchronous Hidden Markov Model (AHMM) to model the relationship between people. Furthermore, we propose a group activity detection algorithm which can handle both symmetric and asymmetric group activities, and demonstrate that this approach enables the detection of hierarchical interactions between people. Experimental results show the effectiveness of our approach.

## I. Introduction

Detecting human group behavior or human interactions has attracted increasing research interests [1-6]. Some example group events of interests include people fighting, people being followed, people walking together, terrorist launching attacks in groups, etc. Being able to automatically detect group activities of interests is important for many security applications. In this paper, we address the following issues for group event detection.

### *A. Group Event Detection with a Varying Number of Group Members*

Most previous group event detection researches [1-2] use a Hidden Markov Model (HMM) or its variation to model the human interactions. Some researchers try to recognize human interactions based on a content-independent semantic set [3-4]. However, most of these works are designed to recognize group activities with a fixed number of group members, where the input feature vector length is fixed. They cannot handle the case where the number of group members is varying, which often occurs in our daily life (e.g., people may leave or join a group activity). In this case, the input feature vector length may vary with different number of group members. Although some works [5-6] tried to deal with the detection of group activities with a varying number of members, most of them have assumptions under some specific scenarios which restrict their applications.

### *B. Group Event Detection with a Hierarchical Activity Structure*

In many scenarios, interacting people form subgroups. However, these subgroups are not independent to each other and they may further interact to form a hierarchical structure. For example, in Fig. 1, three people fighting form a subgroup of *fighting* (the dashed circle). At the same time, another person is approaching the three fighting people and these four people form a larger group of *approaching* (the solid circle in Fig. 1). This is an example of hierarchical activity structure with the group of *approaching* at a higher level than the group of *fighting*. Some algorithms [1-2] could be extended to deal with the problem of hierarchical structure event detection when the number of group members is fixed. Our work addresses the problem of group event detection with *a varying number of group members* under a *hierarchical activity structure*.

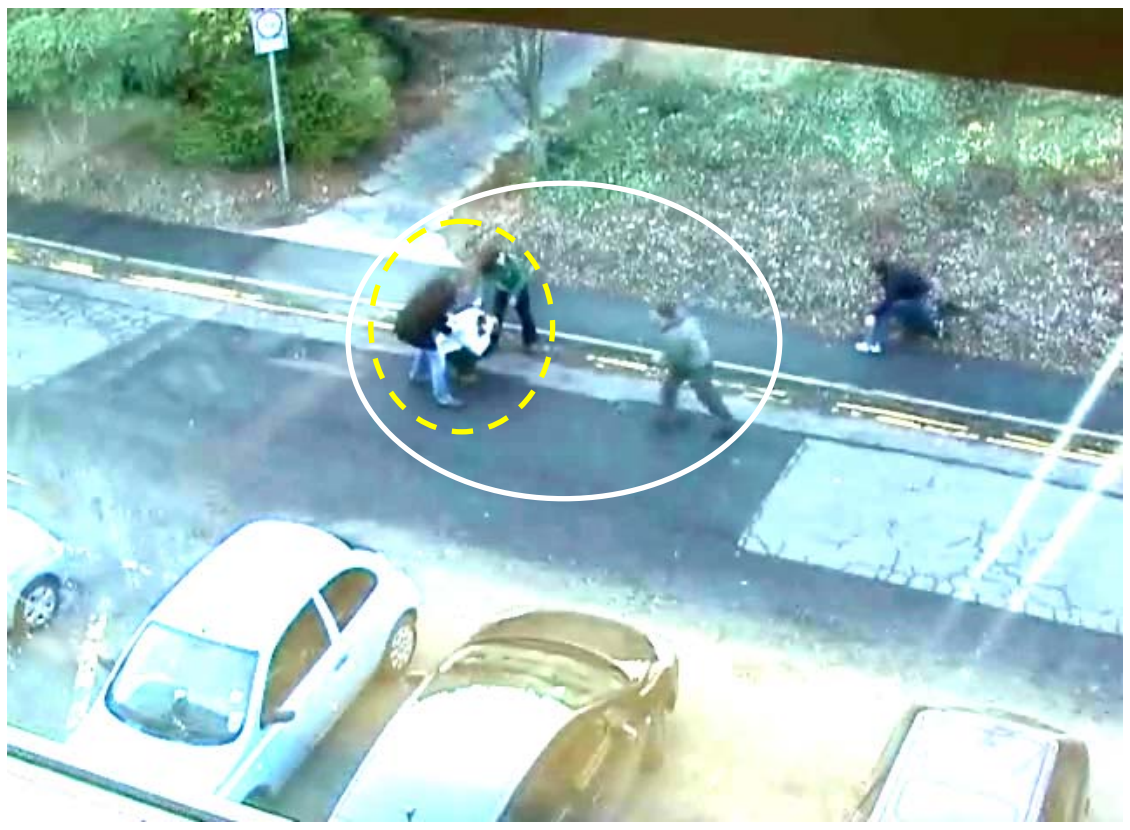

Fig. 1. Example of a group event with hierarchical activity structure [8].

### *C. Clustering with an Asymmetric Distance Metric*

Most previous clustering algorithms [6,10] perform clustering based on a symmetric distance metric (i.e., the distance between two people is symmetric regardless of the relationship of the people).

In the group event detection, some activities such as “following” are asymmetric (e.g. “person $i$ following person $j$” is not the same as “person $j$ following person $i$”). Defining a suitable asymmetric distance metric and performing clustering under the asymmetric distance metric is an important issue.

The contributions of this paper are summarized as follows:

1) To address the problem of detection with a hierarchical activity structure, we propose a Symmetric-Asymmetric Activity Structure (SAAS).
2) To address the problem of detecting events with a varying number of people, we propose to use a Group Representative (GR) to represent each symmetric activity sub-group.
3) To address the problem of clustering with an asymmetric distance metric, we propose a Seed-Representative-Centered clustering algorithm (SRC clustering) to cluster people with asymmetric distance metric. We combine these contributions into a Group-Representative-based Activity Detection (GRAD) algorithm.

The rest of the paper is organized as follows. Section II describes the distance metric for modeling the activity correlation between two people, which is used in our SRC clustering. Section III describes the proposed SAAS. Section IV describes the SRC clustering algorithm. Section V describes the definition of group representative and its use in the GRAD algorithm. Experimental results are shown in Section VI. Section VII discusses some possible extensions of the algorithm. We conclude the paper in Section VIII.

## II. The Activity Correlation Metric Between People

In this paper, we use the Asynchronous Hidden Markov Model (AHMM) [1,7] to model the activity correlation metric between two people. It should be noted that our proposed GRAD algorithm, as to be detailed later, is general and can easily be extended to use other models [2,12,13,14].

AHMM was introduced to handle asynchronous feature streams. As in Fig. 2, assume there are two asynchronous observation (or feature) sequences $F_i(1:S)$ for person $i$ from time $1$ till time $S$ and $F_j(1:T)$ for person $j$ from time $1$ till time $T$ with the length $T \geq S$, the AHMM tries to associate the corresponding features in order to obtain a better match between streams.

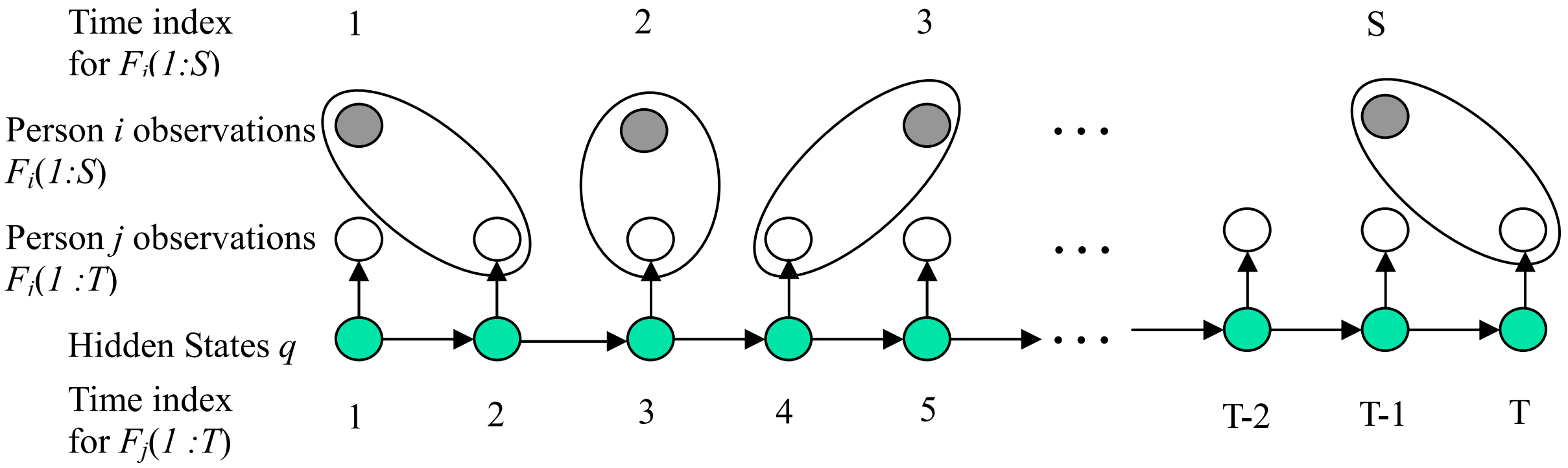

Fig. 2. The Asynchronous Hidden Markov Model (AHMM) for the observation of independent individuals *i* and *j* over the time periods *1:S* and *1:T*, respectively.

The probability that the system emits the next observation of sequence $F_i$ at time $t$ while in state $q(t)=k$, as defined in [7], is,

$$\varepsilon(k,t) = P\left(\tau(t) = s \middle| \tau(t-1) = s-1, q(t) = k, F_i(1:s), F_j(1:t)\right) \tag{1}$$

where $P(\cdot)$ represents the probability. The additional hidden variable $\tau(t) = s$ can be seen as the alignment between $F_i$ and $q$ (and $F_j$ which is also aligned with $q$). Based on Eqn (1), we can define the forward procedure as in Eqn (2) [7].

$$\alpha(s,k,t) = p\left(\tau(t) = s, q(t) = k, F_i(1:s), F_j(1:t)\right) \tag{2}$$

$$= \varepsilon(k,t) p(F_i(s), F_j(t) \mid q(t) = k) \sum_{l=1}^{N} P(q(t) = k \mid q(t-1) = l) \alpha(s-1, l, t-1)$$

$$+ (1 - \varepsilon(k,t)) p(F_j(t) \mid q(t) = k) \sum_{l=1}^{N} P(q(t) = k \mid q(t-1) = l) \alpha(s, l, t-1)$$

where $p(\cdot)$ represents the distribution, $F_i(s)$ and $F_j(t)$ are the observations for persons $i$ and $j$ at time $s$ and $t$, respectively, and $N$ is the total number of hidden states.

Therefore, the activity correlation metric for person $j$ with respect to person $i$ under activity $\theta$ at time $t$ can be calculated as:

$$co_i^{\theta}(j,t) = \sum_{k\in\theta} P(q(t)=k \mid F_i(1:s), F_j(1:t)) = \frac{\sum_{k\in\theta}\sum_{s=t-\Delta t}^{s=t+\Delta t}\alpha(s,k,t)}{\sum_{k=1}^{N_s}\sum_{s=t-\Delta t}^{s=t+\Delta t}\alpha(s,k,t)} \qquad (3)$$

where $k \in \theta$ means all the states that belong to the model of activity $\theta$. $N_s$ is the total number of states over all activities.

We call the activity with the largest $co_i^{\theta}(j,t)$ *the label for j with respect to i* ( $L_i(j)$ ), which is defined in Eqn (4):

$$L_i(j) = \max_{\theta} co_i^{\theta}(j,t) \qquad (4)$$

The reason of using AHMM for modeling the activity correlation metric is that AHMM can handle asynchronous feature streams. Since the feature streams of different people in the same group may not be perfectly synchronized (e.g., when two people walk together, one person may stretch the leg earlier than the other person), AHMM can help reduce the possible recognition errors from these action asynchronies, as will be demonstrated in the experimental results.

Also, from Eqn (3) and (4), we can see that the activity correlation metric is not symmetric (e.g., $co_i^{\theta}(j,t) = \sum_{k\in\theta} P(q(t)=k \mid F_i(1:s), F_j(1:t)) \neq co_j^{\theta}(i,t) = \sum_{k\in\theta} P(q(t)=k \mid F_j(1:s), F_i(1:t))$ because the order of $F_i$ and $F_j$ has been changed, and similarly $L_i(j)$ may not equal to $L_j(i)$). Therefore, when we use this activity correlation metric as the distance metric for clustering, we need to deal with the problem of clustering with an asymmetric distance metric as will be described in detail in Section IV.

## III. Symmetric and Asymmetric Activities

To solve the problem of the hierarchical activity structure, we classify activities into symmetric activities and asymmetric activities. Assume we have two entities *i* and *j*, the activity $\theta$ between *i* and *j* is defined as a symmetric activity if "*i* has the activity $\theta$ with *j*" is the same as "*j* has the activity $\theta$ with *i*". For example, the activity *WalkTogether* is a symmetric activity because "*i walking together with j*" is the same as "*j walking together with i*". From the above definition, we see that entities belonging to the same symmetric activity play similar roles for the activity and are interchangeable. We can further define the symmetric group as a group of entities where any two entities in the group perform the same symmetric activity. A symmetric group can have a varying number of group members or entities. It should be noted that we also extend the definition of symmetric group to include single entity activity cases. For example, if a person walks alone and does not have any symmetric activity interaction with other people, this single person can form a symmetric group of *walking*.

Similarly, the activity $\theta$ between $i$ and $j$ is defined as an asymmetric activity if the activity is not a symmetric activity. For example, the activity *Following* is an asymmetric activity because "*i* is *following j*" is different from "*j* is *following i*".

With the introduction of symmetric activity and asymmetric activity, we proposed to solve the hierarchical-activity-recognition problem by first clustering people into non-overlapping symmetric groups and then modeling the asymmetric-activity interactions between the symmetric groups. We call this the Symmetric-Asymmetric Activity Structure (SAAS). For example, in the example of Fig. 1, we can first cluster people into two symmetric groups: the three-people fighting group and one person walking group. Then the asymmetric activity *Approaching* between these four people can be modeled as the interaction between the fighting group and the walking group. It should be noted that the idea of the proposed SAAS is general and can easily be extended to model other hierarchical activity structures. For example, we can model the symmetric activities of two *WalkTogether* groups as the lower level activity and model the symmetric activity *Ignore* (i.e. people ignore each other) between these two groups as the higher level activity, thus forms a Symmetric-Symmetric Activity Structure (SSAS).

## IV. The SRC clustering algorithm

Based on the description of SAAS, before detecting the symmetric activity of each symmetric group and the asymmetric activity between symmetric groups, we need to cluster people into symmetric groups first. In this section, we propose a Seed-Representative-Centered clustering (SRC clustering) algorithm. The algorithm is described as follows:

Step 1) Detecting the cluster seeds. Two kinds of cluster seeds are defined.

1) *Active people in the group.* Person *i* will be considered as an *active* person in the group if

$$C_i(t) > T_C \tag{5}$$

where $C_i(t)$ is the change of body size of person *i* at time *t* and $T_c$ is a threshold. $C_i(t)$ is calculated by $C_i(t) = \left|W_i(t) \cdot H_i(t) - W_i(t-1) \cdot H_i(t-1)\right| / \left(W_i(t) \cdot H_i(t)\right)$ where $W_i(t)$ and $H_i(t)$ are the width and height of the Minimum Bounding Box (which is the smallest rectangular box that includes the person in motion [9]) of person *i* at time *t*.

2) *The people pairs with high activity correlation metric values.* People pairs *i* and *j* with high activity correlation metric values will also be considered as cluster seeds, if

$$\begin{cases} co_i^L(j,t) > T_o \text{ and } co_j^L(i,t) > T_o \\ L_i(j) = L_j(i), \text{ and} \\ L_i(j) \text{ is a symmetric activity} \end{cases} \tag{6}$$

where $T_o$ is a threshold to decide where people pairs *i* and *j* have high activity correlation.

Step 2) Post-processing of the cluster seeds. After detecting the cluster seeds, a post processing is performed to combine seeds that belong to the same symmetric group. Cluster seeds with the same symmetric activity label will be combined together. For example, if (*a*,*b*) is a cluster seed and *c* is another cluster seed, *c* can be combined with (*a*,*b*) to form a larger seed of (*a*,*b*,*c*) if $L_a(b)=L_a(c)=L_c(a)$.

Step 3) Calculate Seed Representatives (SR) for the cluster seeds. We can combine people in the same cluster seed to create a Seed Representative (SR) for each cluster seed. There can be many ways to define the Seed Representative. For example, we could pick any feature vector close to the cluster center as the Seed Representative. In this paper, the average feature vector of people in the same seed is used as the SR for the cluster seeds.

Step 4) Cluster the remaining people based on the SRs. The calculated Seed Representatives serve as the centers of clusters and the remaining people are clustered around them. A person $i$ is grouped into the cluster indicated by the SR $K$ if $co_i^L(K,t)$ is maximum and $L_i(K)$ is a symmetric activity. It should be noted that only the Seed-Representative-Centered (SR-Centered) metric value is used for clustering in this step. The SR-Centered metric value is defined as:

$co_i^L(K,t)$ is an SR-Centered metric value if K is a SR and i is not a SR.

Since only the SR-Centered metric value is used for clustering, the asymmetry problem of the activity correlation metric is avoided.

As a summary, the proposed SRC clustering algorithm extracts only high correlation pairs as well as single active person in the seed detection step and use only the SR-Centered value in the clustering step. Therefore, it can deal with the problem of clustering with an asymmetric distance metric.

## V. Group Representative and the GRAD algorithm

### *A. The Definition of Group Representative*

As mentioned, people in the same symmetric group are interchangeable and play a similar role. Based on this property, each symmetric group can be represented by a single entity, which we call the Group Representative (GR). There can be different ways to define the GR. In this paper, we investigate three ways to define the GR. They are described as follows:

1) **Physical GR (P-GR).** The Physical Group Representative is an actual person selected from the symmetric group. We define P-GR as the most representative person of the symmetric group which has the highest joint value for representing the group's activity $\theta_A$ as well as correlating with other people in the symmetric group, as in Eqn (7).

$$P\text{-}GR_A(t) = \max_i \big(p(F_i(t) \mid \theta_A) \cdot p_0(i,\theta_A,t)\big) \tag{7}$$

where $P\text{-}GR_A(t)$ is the P-GR for symmetric group $A$ at time $t$, $F_i(t)$ is the feature vector of person $i$ at time $t$, $\theta_A$ is the activity for $A$, and $p_0(i,\theta_A,t) = exp\left(\sum_{\substack{j\in A \\ and\ j\neq i}} co_j^{\theta_A}(i,t)\right)$. In Eqn (7), $p(F_i(t) \mid \theta_A)$ reflects the representativeness of person $i$ for activity $\theta_A$, and $p_0(i,\theta_A,t)$ can be viewed as a prior which measures the distance or correlation of person $i$ to other people in $A$ [11].

2) **Virtual GR (V-GR).** The virtual GR is not an actual person. Rather, it is the combination of multiple people in the same symmetric group. The V-GR is defined as the average of *all* people in the feature space in the same symmetric group. Therefore, the feature vector of V-GR at time $t$ can be defined as:

$$F_{V-GR_A}(t) = \underset{i\in A}{avg}(F_i(t)) \tag{8}$$

where $F_i(t)$ is the feature vector for person $i$ at time $t$, and group $A$ is the symmetric group.

3) **Selective Virtual GR (SV-GR).** Similar to V-GR, SV-GR is also a virtual GR which is the combination of multiple people. However, SV-GR is the average of only those most representative people for the symmetric group, as in Eqn (9).

$$F_{SV-GR_A}(t) = \underset{i\in R_A}{avg}(F_i(t)) \tag{9}$$

where $F_{SV-GR_A}(t)$ is the feature vector of SV-GR for group $A$ at time $t$, $F_i(t)$ is the feature vector for person $i$ at time $t$. $R_A = \{i \mid N\big(p(F_i(t) \mid \theta_A) \cdot p_0(i,\theta_A,t)\big) > T_R\}$, where $T_R$ is a threshold to decide whether person $i$ is representative. $N(\cdot)$ is the normalization operation such that $\sum_i N\big(p(F_i(t) \mid \theta_A) \cdot p_0(i,\theta_A,t)\big) = 1$.

### *B. The GRAD Algorithm*

With the introduction of GR as well as our proposed SAAS and SRC clustering algorithm, we propose a Group-Representative-based Activity Detection (GRAD) algorithm to solve the problem of detecting group events with a varying number of group members under a hierarchical activity structure. The GRAD algorithm can be summarized as follows:

Step 1) For each frame *t*, people are first clustered into non-overlapping symmetric groups by the SRC clustering algorithm (the dotted ellipses in Fig. 3). The symmetric activity for each symmetric group can then be recognized. In this paper, we propose the following two methods to recognize the symmetric activity.

1) Directly use the activity label for each cluster seed as the recognized activity for the symmetric group.

2) A more sophisticated way is to extract some group features [5,15] from the symmetric group and use a separate model such as HMM for recognition, as described by Eqn (10).

$$\theta_A(t) = \max_{\theta}\left(p(F_A(t) \mid \theta) \cdot p_1(\theta,t)\right) \tag{10}$$

where $p_1(\theta,t) = \exp\left(\sum_{i,j \in A} co_i^{\theta}(j,t)\right)$ can be viewed as a prior for activity [11]. $F_A(t)$ is the global feature vector for symmetric group *A*, and $p(F_A(t) \mid \theta)$ is the probability calculated by the model used for recognizing symmetric activities.

Step 2) Each symmetric group is represented by a Group Representative (the two bold solid circles in Fig. 3).

Step 3) The asymmetric activity between symmetric groups is then captured by the interaction of the GR of each symmetric group (the bold solid line in Fig. 3). In this paper, we detect the asymmetric activity between two symmetric groups based on the activity correlation metric between GRs, as in Eqn (11).

$$\theta_{A,B}(t) = \max_{\theta}\left(co_{GR_B}^{\theta}(GR_A,t) \cdot p_2(\theta,t)\right) \tag{11}$$

where $p_2(\theta,t) = \exp\left(\sum_{i\in A, j\in B} co_j^{\theta}(i,t)\right)$ is the prior for asymmetric activity $\theta$. *A* and *B* are two symmetric groups. Since the activity correlation metrics are not symmetric, in our notations, we put the GRs in the order according to a specific feature such as the average speed of the symmetric group (i.e. the average speed for group *A* is smaller than *B* in $co_{GR_B}^{\theta}(GR_A)$). Furthermore, as mentioned, the activity between two symmetric groups can also be symmetric (e.g. two groups *Ignore* each other). In this case, the interaction of the GR can also be used to detect the symmetric activity between two groups.

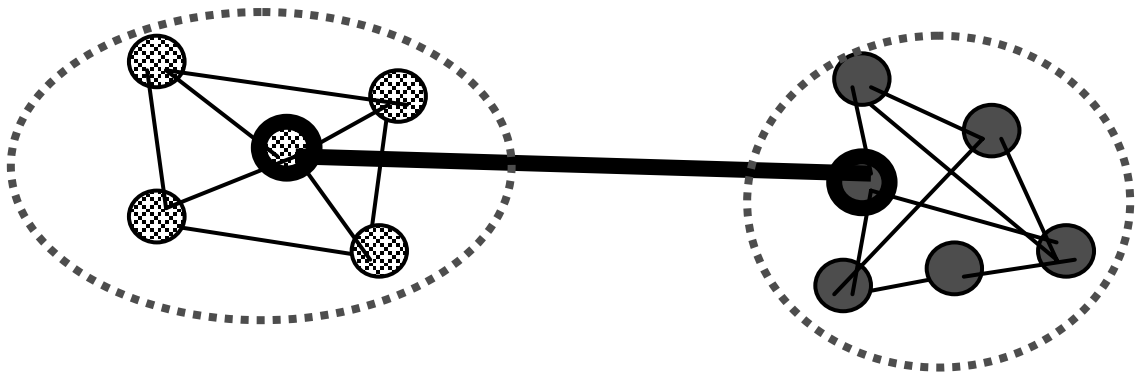

Fig. 3. The GRAD algorithm.

In the GRAD algorithm described above, since we use a single GR to represent each symmetric group, we always have a fixed input feature vector length. Therefore, we can solve the problem of group event detection with a varying number of group members.

### *C. Discussion of Group Representative*

Since we have all the activity correlation metrics between any two people, there can be alternative methods to deal with the detection-with-a-varying-number-of-members problem. For example, we can use the Majority Vote method (MV) [17,18] for asymmetric activity recognition by taking the majority vote from all the asymmetric activity labels between people pairs from two symmetric groups as the resulting activity label. Compared with MV and other methods, the major difference of our proposed GR method is to use a single representative (physical or virtual) to represent the whole symmetric group. With the introduction of GR, we can have the following advantages:

1) Methods such as MV lack a global view of the whole group since all the activity correlation

metrics only reflect the local information between two people. However, when selecting the GR by Eqn (7)-(9), we are actually checking the whole symmetric group. Therefore, the selected GR will have a global view of the whole group.

2) More importantly, when detecting the asymmetric activity between two symmetric groups, some people that are not highly related may disturb the recognition result. For example, as in Fig. 4, the asymmetric activity $\theta$ between $A$ and $B$ is mainly decided by the interaction between the bold-faced people (i.e. bold-faced circles in Fig. 4) in $A$ and the bold-faced people in $B$. The dotted person located on the side of $A$ does not have high correlation in $\theta$ with people in $B$ and may have misclassified activity label with $B$. This dotted person is an outlier and may disturb the recognition results. When using methods such as MV to perform recognition, the dotted outlier person is included and the recognition accuracy may be decreased. However, when using GR with our proposed method (especially the P-GR and the SV-GR), the low-correlated outlier person will be discarded from the asymmetric activity detection process, thus reducing the disturbance from these outlier people. Therefore, our proposed GR will also increase the recognition accuracy by efficiently discarding outliers.

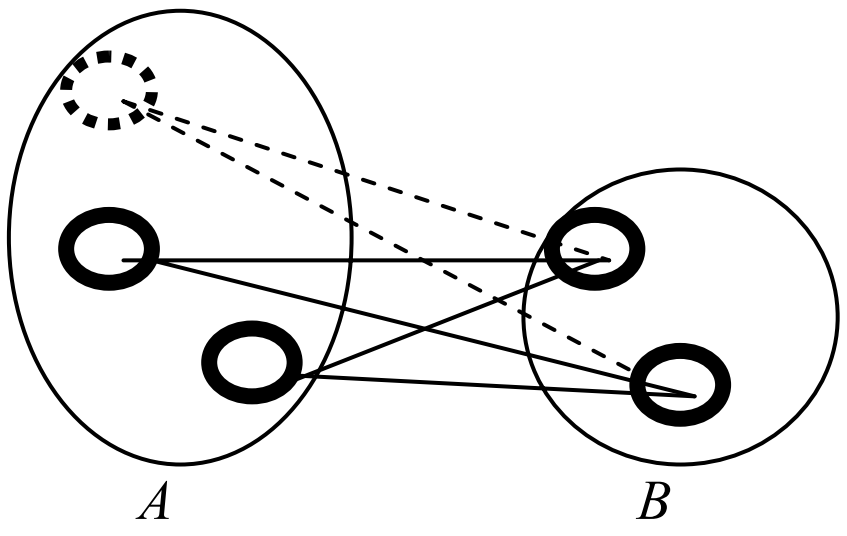

Fig. 4. An example of the disturbance from an outlier person (dotted circle: outlier person, bold-faced circle: regular person).

## VI. Experimental Results

In this section, we show experimental results for our proposed methods and compare our results with other methods. We perform experiments based on the BEHAVE dataset [8]. Six long sequences are selected in our experiments with each sequence including 7000 to 11000 frames. We try to detect eight group activities: *InGroup, Approach, WalkTogether, Split, Ignore, Chase, Fight, RunTogether*.

Some example video frames are shown in Fig. 5. The definitions of these eight activities are listed in Table 1. We classify these eight activities into two categories with *InGroup, WalkTogether, Ignore, Fight* and *RunTogether* as symmetric activities, and *Approach, Split and Chase* as asymmetric activities. It should be noted that we extended the definition of activity *Ignore*. The two people will *ignore* each other if they do not have other activity correlation. Furthermore, *Ignore* will also be used to model the non-interaction case between two symmetric groups. We also add a *single* activity into the symmetric activity list for those people that cannot be clustered into any symmetric group.

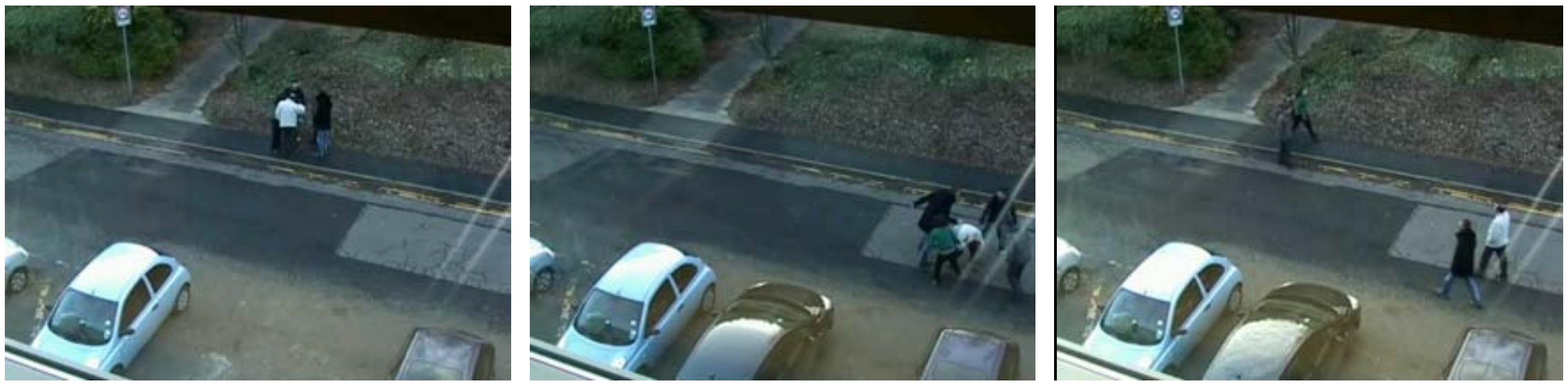

InGroup Fight WalkTogether

Fig. 5. Some example video frames for the group activities [8].

Table 1 The definition of group activities
(Activities in grey are symmetric activities and activities in white are asymmetric activities)

| **Activity** | **Definition** |
|---|---|
| *InGroup* | The people are in a group and not moving very much |
| *WalkTogether* | People walking together |
| *Fight* | Two or more groups fighting |
| *RunTogether* | The group is running together |
| *Ignore* | Ignoring of one another |
| *Approach* | Two people or groups with one (or both) approaching the other |
| *Split* | Two or more people splitting from one another |
| *Chase* | One group chasing another |

For simplicity, we only use the Minimum Bounding Box (MBB) information [9] to derive all the features used for group activity recognition. Note that the proposed algorithm is not limited to the MBB features. Other more sophisticated features [19,20] can easily be applied to our algorithm to give better results. Six features are used for calculating the persons' activity correlation metrics in Eqn (3). They are listed in Table 2.

Table 2 The definition of input features

| Feature Name | Definition |
|---|---|
| Change of Width | $\left\| W_i(t) - W_i(t-1) \right\| / W_i(t)$ |
| Change of Height | $\left\| H_i(t) - H_i(t-1) \right\| / H_i(t)$ |
| Speed | $\sqrt{\left(x_i(t) - x_i(t-1)\right)^2 + \left(y_i(t) - y_i(t-1)\right)^2}$ |
| Average Distance | $\sqrt{\left(x_i(t) - \frac{x_i(t) + x_j(t)}{2}\right)^2 + \left(y_i(t) - \frac{y_i(t) + y_j(t)}{2}\right)^2}$ |
| Speed Difference | $\frac{\sqrt{\left(x_i(t) - x_i(t-1)\right)^2 + \left(y_i(t) - y_i(t-1)\right)^2} - \sqrt{\left(x_j(t) - x_j(t-1)\right)^2 + \left(y_j(t) - y_j(t-1)\right)^2}}{2}$ |
| Motion Direction Angle | $\arctan\left(\frac{y_i(t) - y_i(t-1)}{x_i(t) - x_i(t-1)}\right) - \arctan\left(\frac{y_j(t) - y_i(t)}{x_j(t) - x_i(t)}\right)$ |
| Note: The features in this table forms an input feature vector for *i* when calculating its correlation with *j*. $(x_i(t), y_i(t))$ is the center of MBB for *i* at time *t*. $W_i(t)$ and $H_i(t)$ is the width and height of the MBB for *i* at time *t*, respectively. | |

In order to exclude the effect of the tracking algorithm, we use the ground-truth tracking data which is available in the BEHAVE dataset to get the MBB information. In practice, various practical tracking methods [15,21] can be used to obtain the MBB information. Furthermore, the thresholds $T_c$, $T_o$ and $T_R$ in Eqn. (5), (6) and (9) are set to be *0.1*, *0.95* and *0.3*, respectively. These values are manually selected based on the statistics from one of the training sets. In practice, these thresholds can also be selected by some more sophisticated ways such as the validation set method [9].

In our experiments, four methods are compared. For all the HMMs or AHMMs in these methods, we use two hidden states for each activity (plus the starting state and the finishing state, there are in total four states) and a two-mixture Gaussian Mixture Model (GMM) [23,24] for modeling the emission probability for each hidden state. It should be noted that the methods selected to compare in our experiments are typical and the results can easily be extended to other related methods [2,13,16,19]. The four methods are described as follows:

1) **HMM.** Use a single HMM [12,21] to recognize either the symmetric activites or the asymmetric activities. When recognizing symmetric activities, the group features in Table 3 are calculated for each symmetric group. However, it should be noted that the traditional HMM cannot deal with the recognition of hierarchical-structure activities (i.e., a single HMM cannot recognize a lower-level symmetric activity and an upper-level asymmetric activity at the same time). Furthermore, since the input feature vector length is fixed, it also cannot recognize activity with varying number of group members.

Table 3 The definition of group features

| Feature Name | Definition |
|---|---|
| Average Change of Width | $\sum_{i\in A}\left(\lvert W_i(t)-W_i(t-1)\rvert / W_i(t)\right) / \sum_{i\in A} 1$ |
| Average Change of Height | $\sum_{i\in A}\left(\lvert H_i(t)-H_i(t-1)\rvert / H_i(t)\right) / \sum_{i\in A} 1$ |
| Average Speed | $\sum_{i\in A}\sqrt{(x_i(t)-x_i(t-1))^2+(y_i(t)-y_i(t-1))^2} / \sum_{i\in A} 1$ |
| Average Distance | $\sum_{i\in A}\left(\sqrt{\left(x_i(t)-\sum_{j\in A}x_j(t) / \sum_{j\in A}1\right)^2+\left(y_i(t)-\sum_{j\in A}y_j(t) / \sum_{j\in A}1\right)^2}\right) / \sum_{i\in A} 1$ |
| Speed variance | $\sum_{i\in A}\left(\sqrt{(x_i(t)-x_i(t-1))^2+(y_i(t)-y_i(t-1))^2}-average_speed\right)^2 / \sum_{i\in A} 1$ |
| Note: the definition of $x_i(t)$, $y_i(t)$, $W_i(t)$ and $H_i(t)$ are the same as in Table 2. $A$ is a symmetric group. | |

2) **Layered HMM+SAAS.** In [1], a layered HMM is proposed. In our experiment, we extend this layered HMM based on our proposed SAAS to recognize hierarchical-structure group activities, where the HMMs in the lower layer recognize the symmetric activities for each symmetric sub-group and the HMM in the higher layer takes the outputs of the lower layer as input to recognize asymmetric activities, as in Fig. 6. Furthermore, extra features are also calculated as input to the higher layer HMM [1]. In our experiment, we use hard decision outputs [1] of the lower layer HMMs as the input to the higher layer HMM. Furthermore, features in Table 2 are used as the extra features for inputting to the higher layer HMM. The extra features are calculated between two symmetric sub-groups. However, similar to HMM, since the input feature vector length of the layered HMM is also fixed, it cannot deal with the problem of activity recognition with varying number of group members.

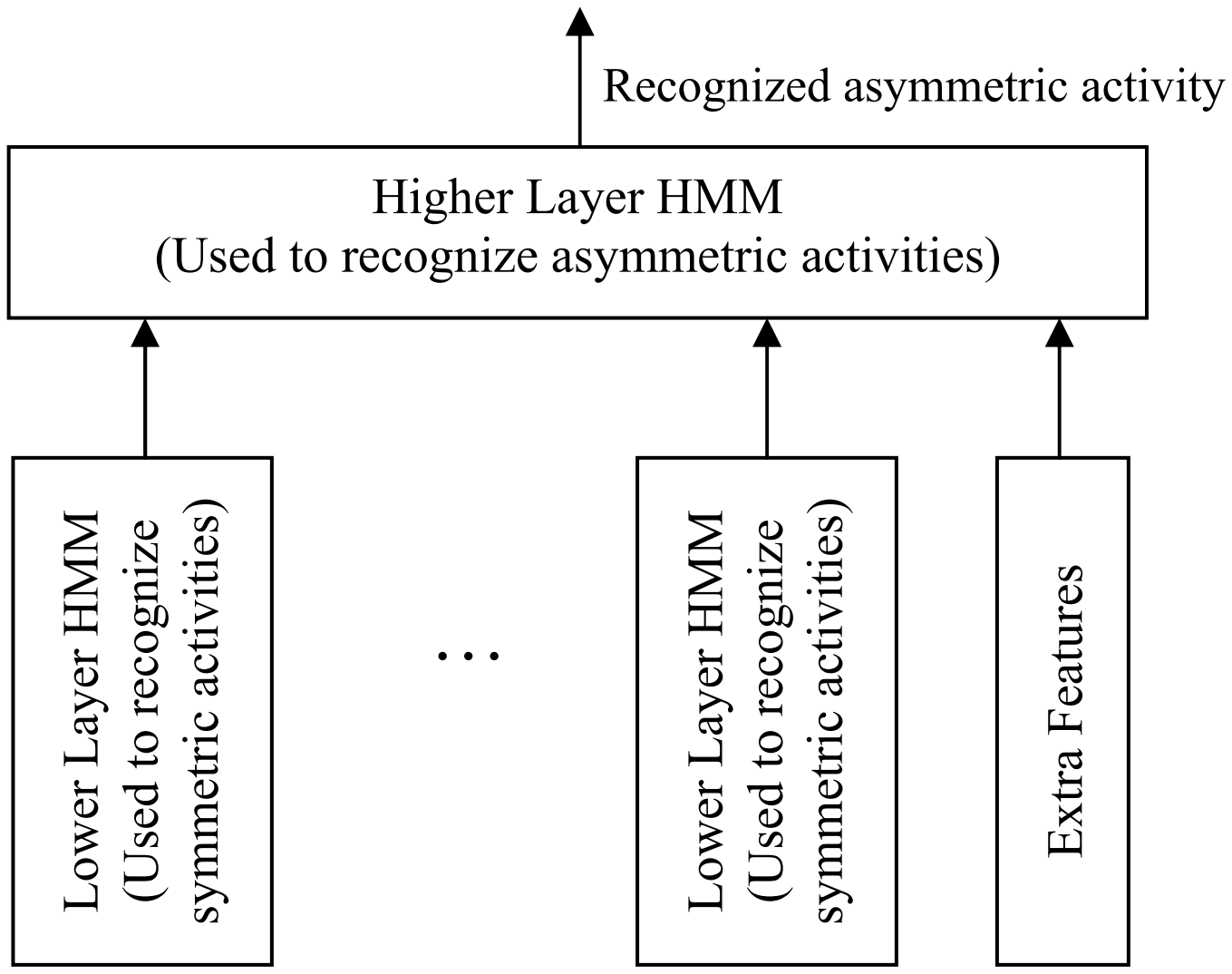

Fig. 6. The Layered HMM (Lower layer HMMs are used to recognize symmetric activities and a higher layer HMM is used to recognize asymmetric activities).

3) **SAAS+SRC+MV.** Based on the proposed SAAS, it uses our proposed SRC clustering algorithm to cluster people into symmetric groups and detect the activity of these symmetric groups, then uses the Majority Vote to detect the asymmetric activities between the symmetric groups. When detecting the symmetric activities, two different methods are used: (a) use the activity label for each cluster seed as the recognized activity for the symmetric group (SAAS+SRC+MV-1 in Tables 5-9), and (b) calculate the group features from the symmetric group and use the HMM model for recognition, as in Eqn (10) (SAAS+SRC+MV-2 in Tables 5-9). The SAAS+SRC+MV method can recognize hierarchical-structure activities as well as activities with a varying number of group members. However, it should be noted that using only Majority Vote cannot recognize hierarchical-structure activities and varying-member activities. By combining MV with our proposed SAAS and SRC clustering algorithm, it can deal with these activities.

4) **The GRAD algorithm (SAAS+SRC+GR).** Use the GRAD algorithm to detect group activities. Use our proposed SAAS and SRC clustering to cluster people and detect symmetric activities. However, different from the SAAS+SRC+MV method which uses MV to detect asymmetric activities, the GRAD algorithm uses our proposed GR to detect asymmetric activities. Similar to the SAAS+SRC+MV method, we use two different methods to detect symmetric activities. They are: (a) use the cluster seed label as the recognized activity (GRAD-1 in Tables 5–9), and (b) use an independent HMM to recognize the symmetric activities (GRAD-2 in Tables 5–9).

Experiments for four scenarios are designed to compare the four methods described above, they are (a) recognizing *hierarchical-structure activities* with a *varying number* of group members (*hierarchical+varying* in Table 4), (b) recognizing *only symmetric activities* with *fixed number* of group members (*symmetric+fixed* in Table 4), (c) recognizing *only asymmetric activities* with *fixed number* of group members (*asymmetric+fixed* in Table 4), and (d) recognizing *hierarchical –structure activities* with *fixed number* of group members (*hierarchical+fixed* in Table 4). These four sets of experiments will be described in detail in the following sections. Table 4 summarizes the capabilities of the four methods in dealing with these four experimental tasks. It should be noted that the scenario of *hierarchical+varying* is the general case for group activities and the other scenarios can be viewed as the special cases for this scenario.

Table 4 The capabilities of different methods in dealing with different experimental tasks (Note: label "○" means the method is able to deal with the corresponding task, the label "×" means the method is unable to deal with the corresponding task)

| | HMM | Layered-HMM | SAAS+SRC+MV | GRAD (SAAS+SRC+GR) |
|---|---|---|---|---|
| symmetric+fixed | ○ | ○ | ○ | ○ |
| asymmetric+fixed | ○ | ○ | ○ | ○ |
| hierarchical+fixed | × | ○ | ○ | ○ |
| hierarchical+varying | × | × | ○ | ○ |

***A. Experimental results for recognizing only symmetric activities with a fixed number of group members***

In this section, we compare the performances of the four methods in recognizing only the four symmetric activities (i.e. *InGroup, WalkTogether, Fight* and *RunTogether*). Furthermore, we assume that the symmetric groups have already been clustered and the number of members in all symmetric groups is fixed to 3. In order to fix the member for all groups to 3, we discard the activity segments from the dataset whose group members are less than 3. For activity segments with more than 3 members, we manually pick three members to form a symmetric group.

We perform experiments under 50% training and 50% testing. Five independent experiments are performed and the results are averaged. The experimental results are listed in Table 5. In Table 5, the Total Frame Error Rate (TFER) [9,25] is compared. TFER is defined by $N_{t_miss} / N_{t_f}$, where $N_{t_miss}$ is the total number of misdetection frames for all activities, and $N_{t_f}$ is the total number of frames in the test set. TFER reflects the overall performance of each algorithm in recognizing all these five symmetric activities.

Table 5 TFER comparison for symmetric activity recognition with fixed number of group members

| Methods | TFER |
|---|---|
| Set-1<br>(HMM, Layered HMM+SAAS, SAAS+SRC+MV-2 and GRAD-2) | 5.36% |
| Set-2<br>(SAAS+SRC+MV-1 and GRAD-1) | 5.52% |

Since only symmetric activities with a fixed number of people are recognized in this experiment, the HMM method, the Layered HMM+SAAS method, the SAAS+SRC+MV-2 method, and the GRAD-2 method are exactly the same to each other and they can be classified as one set (Set-1 in Table 5). Similarly, the SAAS+SRC+MV-1 method and the GRAD-1 method can be classified as another set (Set-2 in Table 5). Basically, the major difference between the methods of these two sets is that methods in Set-1 can have a global view of the whole symmetric group by using the group features, while the methods in Set-2 only use local information of the cluster seeds for recognition.

However, from Table 5, we can see that the TFER for both sets are very close. Similar results can also be found for larger numbers of group members. This implies that since members in the symmetric group are interchangeable and similar, using only local information from parts of the group members may be enough to recognize symmetric activities.

### *B. Experimental results for recognizing only asymmetric activities with a fixed number of group members*

In this section, we perform experiments to recognize the three asymmetric activities (*Approach, Split* and *Chase*). Similar to the previous section, we fixed the number of members in each asymmetric group to 4. We also assume that each asymmetric group contains two symmetric sub-groups with one group containing 3 people and the other group containing 1 person. It should be noted that since the number of group member is fixed in this experiment, the SRC clustering is not needed for the SAAS+SRC+MV method and the GRAD method and thus is skipped.

The TFER result comparison for asymmetric activity recognition under 50% training and 50% testing is shown in Table 6.

Table 6 TFER comparison for asymmetric activity recognition with fixed number of group members

| Methods | TFER |
|---|---|
| HMM | 23.36% |
| Layered HMM+SAAS | 11.75% |
| SAAS+SRC+MV | 14.98% |
| GRAD (SAAS+SRC+GR) | 10.11% |

From Table 6, we have the following observations:

1) The TFER rate for the HMM method is the worst. The main reason is that the HMM method does not differentiate symmetric sub-groups inside the asymmetric group. Instead it directly calculates group features over the whole asymmetric group. This makes it unable to capture the asymmetric interactions between members inside the group. Compared with the HMM method, the other three methods, which perform asymmetric activity recognition based on our proposed SAAS, have better performance. This demonstrates the effectiveness of our SAAS. It should be noted that it is possible to develop better features than the ones in Table 3 to improve the performance of the HMM method for this experiment. However, our SAAS is still important because (a) when the number of group members becomes larger, the interactions between members may be very complicated. It will be very difficult to develop good features for the whole group without considering its lower level structures. (b) In many applications, people are interested in not only the behavior of the whole group but also the behavior of each individual person or sub-groups of people. In this case, the HMM method will require a large number of separate models for each individual person or sub-groups while our SAAS can do all the tasks in one framework.

2) The performance of the GRAD method is better than the SAAS+SRC+MV method. This will be further demonstrated in later experiments.

3) The performance of the GRAD method, which uses P-GR, is slightly better than the Layered HMM+SAAS method. Since we calculate the extra features of the higher layer HMM by taking the average of people in each symmetric sub-group, the Layered HMM+SAAS method can be viewed as an extension of using the V-GR. Therefore, the result further implies that P-GR can improve the results from V-GR by discarding the outliers from recognition. Since both the GRAD method and the Layered HMM+SAAS method can recognize hierarchical structure activities, we will discuss more of these two methods in the following section.

### *C. Experimental results for recognizing hierarchical-structure activities with a fixed number of group members*

In this section, we perform experiments to recognize *hierarchical structure activities* which contain four symmetric activities (*InGroup, WalkTogether, Fight,* and *RunTogether*) and three asymmetric activities (*Approach, Split, and Chase*). Similar to the previous experiment, we fix the number of people in each asymmetric group as 4, and each asymmetric group contains two symmetric sub-groups with one group containing 3 people and the other group containing 1 person. For simplification, we only recognize the symmetric activity of the 3-people sub-group and the asymmetric activity of the 4 people group in this experiment.

As mentioned in Table 4, the HMM method cannot recognize hierarchical-structure activities. Therefore, we only compare the other three methods. Table 7 shows the results for hierarchical-structure activity recognition under 50% training and 50% testing.

Table 7 TFER comparison for *hierarchical-structure activity recognition* with fixed number of group members

| Methods | | TFER | |
|---|---|---|---|
| | | Symmetric Activity | Asymmetric Activity |
| Layered HMM+SAAS | | 5.36% | 11.75% |
| SAAS+SRC+MV | SAAS+SRC+MV-1 | 5.52% | 14.98% |
| | SAAS+SRC+MV-2 | 5.36% | |
| GRAD (SAAS+SRC+GR) | GRAD-1 | 5.52% | 10.11% |
| | GRAD-2 | 5.36% | |

Since the numbers of group members are the same as the previous experiments, the TFER results for symmetric activities and asymmetric activities in Table 7 are exactly the same as those in Table 5 and Table 6, respectively. We can see from Table 7 that the GRAD method and the Layered HMM+SAAS method have similar performance. However, compared with the Layered HMM+SAAS method as well as other HMM-based methods [15-19], our proposed GRAD method has the following advantages:

1) The Layered HMM+SAAS method as well as most other HMM-based methods [2,13,16,19] cannot handle the recognition with a varying number of group members while our GRAD algorithm can handle this problem by the use of the Group Representative.
2) More importantly, there may be hierarchical-structure activities with more than two levels. For example, several asymmetric groups may form a super symmetric group and these super symmetric groups may further form an even larger asymmetric group. In these cases, the HMM-based methods may require very complicated models for recognition which may be very difficult for training and calculation. However, since our GRAD method only extracts GRs from the groups for the recognition in the higher level, it can be kept simple even for those multi-level-structure activities.

### *D. Experimental results for recognizing hierarchical-structure activities with a varying number of group members*

In the above sections, we have demonstrated that our GRAD algorithm has comparable or better results than the previous methods when handling the special scenarios that the previous algorithms can also handle. In this section, we will perform experiments for the general scenario of *hierarchical-structure activities* with a *varying number of group members* and try to recognize all of the group activities in Table 1 for all symmetric and asymmetric groups. From Table 4, we can see that only the SAAS+SRC+MV method and the GRAD method can handle the task in this experiment. Therefore, we only compare these two methods in this section.

In this experiment, we randomly select three long sequences for training and use the other three long sequences for testing. Five independent experiments are performed and the results are averaged.

The experimental results of SAAS+SRC+MV-1 and GRAD-1 are shown in Fig. 7. For the GRAD method, three different GRs are used: (a) Physical GR (P-GR in Fig. 7), (b) Virtual GR (V-GR in Fig. 7) and (c) Selective Virtual GR (SV-GR in Fig. 7). In order to show the advantage of using AHMM, we also includes the results of using regular HMM [22] for modeling the activity correlation metric (with "HMM" in Fig. 7, e.g., SAAS+SRC+MV-1 (HMM)).

In order to take clustering errors into consideration, two error rates are compared in Table 8: the Group Clustering Error Rate (GCER) and the Event Detection Error Rate (EDER). They are defined in Eqn (12) and (13) respectively.

$$GCER = {\#\, of\ clustering\ error\ frames} \big/ {\#\, of\ total\ frames} \tag{12}$$

$$EDER = {\#\, of\ error\ frames} \big/ {\#\, of\ total\ frames} \tag{13}$$

where a frame is a clustering error frame if any person in the frame is mis-clustered into another symmetric group, and a frame is an error frame if any of the following take place: (a) any person in the frame is mis-clustered into another symmetric group, (b) any of the symmetric activities is misclassified, and (c) any of the asymmetric activities is misclassified.

The GCER reflects the performance of the algorithm in clustering people into symmetric groups. The EDER reflects the overall performance of the algorithm in detecting both the symmetric activities and the asymmetric activities.

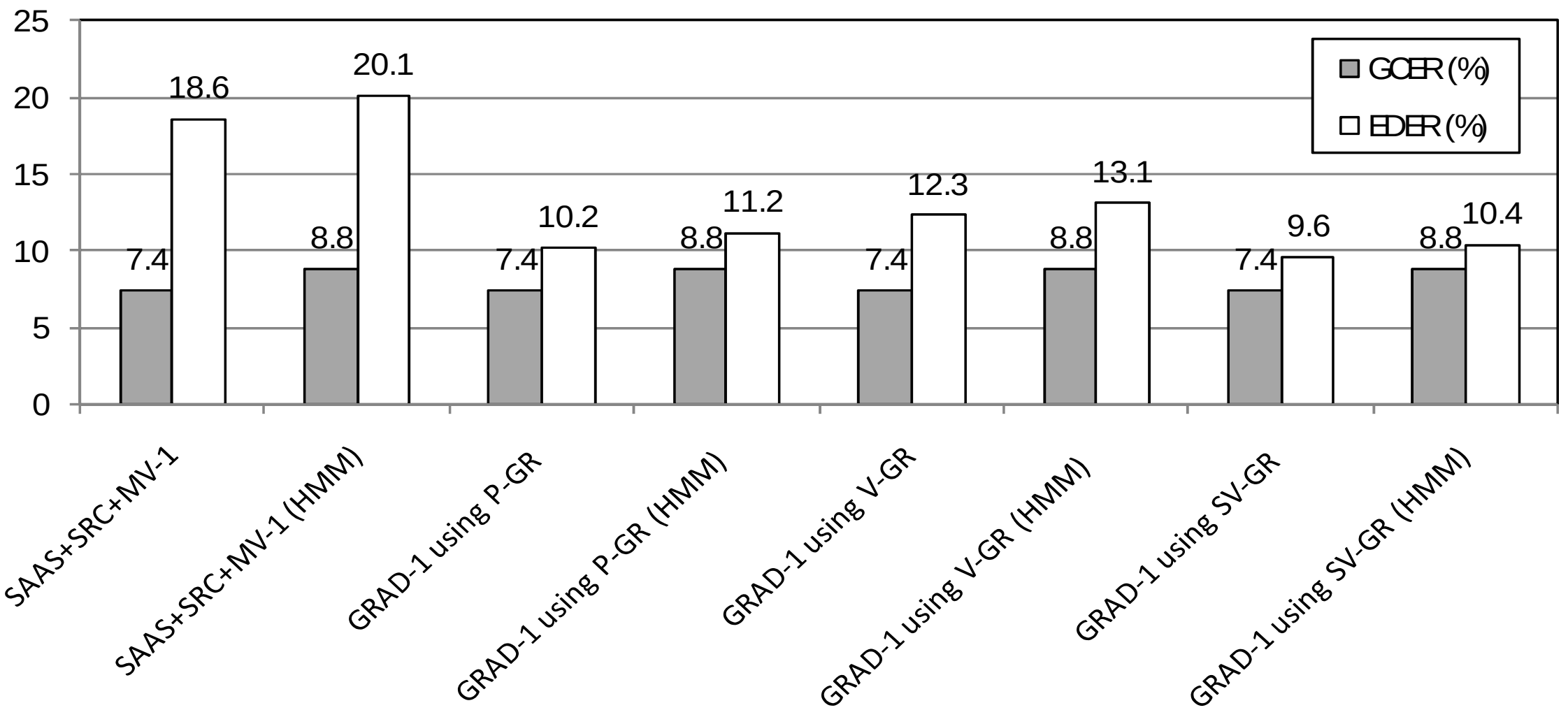

Fig. 7. The experimental results for hierarchical-structure activity recognition with varying number of group members.

Several observations from Fig. 7 are listed below:

Since all methods use the proposed SRC clustering algorithm for clustering people into symmetric groups, their GCERs are the same if using the same activity-correlation-metric model. Therefore, we can see from Fig. 7 that the GCERs of SAAS+SRC+MV-1, GRAD-1 using P-GR, GRAD-1 using V-GR, and GRAD-1 using SV-GR are the same. Similarly, the GCERS of SAAS+SRC+MV-1 (HMM), GRAD-1 using P-GR (HMM), GRAD-1 using V-GR (HMM), and GRAD-1 using SV-GR (HMM) are the same. The low GCER demonstrates the effectiveness of the SRC clustering algorithm. Furthermore, methods using AHMM as an activity-correlation-metric model has a better GCER than those use HMM. This demonstrates that using AHMM can improve the performance by handling the possible action asynchronies.

Comparing the EDER, we can see that the EDERs of the GRAD algorithm are obviously better than that uses majority vote. This supports our claim that the introduction of GR can greatly improve the detection rate for asymmetric activities. Comparing the three GR-based methods, we can see that the EDER of P-GR is better than that of V-GR. This further demonstrates that the P-GR can improve the performance by discarding outliers from asymmetric activity recognition. However, the EDER difference between these two GRs is not large. This is because (a) although V-GR includes outliers, the effect of these outliers is decreased by the averaging with non-outliers, and (b) there may be cases where none of the actual person in the symmetric group is representative enough for the group, in these cases, the P-GR may not perform better than the V-GR. Furthermore, the method using SV-GR has the best EDER. This is because SV-GR has the following two advantages: (a) similar to P-GR, SV-GR can discard outliers by averaging only the most several representative people in the group, and (b) in case when there is no actual person representative for the group, SV-GR can create a virtual GR by averaging several people in the group. However, we can also see from Table 8 that the improvement of SV-GR from P-GR is small. This is because (a) the clustering errors (i.e. GCER) take a large portion of the errors in EDER. This limits the improvement space of SV-GR. It is expected that the performance of the GRAD algorithm can be further improved if people can be clustered better into symmetric groups. (b) Due to the scenarios of the BEHAVE dataset, people in each symmetric sub-group are comparatively close to each other, therefore the chances that none of the actual person is representative are low.

Fig. 8 shows the average False Alarm rate (FA) and Miss Detection rate (Miss) [9] of the GRAD algorithm for the activities in Table 1, where the SV-GR is used for the GRAD algorithm. The Miss rate is defined by $cnt_\theta^{fn} / cnt_\theta^{+}$, where $cnt_\theta^{fn}$ is the number of false negative (misdetection) samples for activity $\theta$, and $cnt_\theta^{+}$ is the total number of positive samples of activity $\theta$ in the test data. The FA rate is defined by $cnt_\theta^{fp} / cnt_\theta^{-}$, where $cnt_\theta^{fp}$ is the number of false positive (false alarm) samples for activity $\theta$, and $cnt_{\theta}^{-}$ is the total number of negative samples of activity $k$ in the test data.

From Fig. 8, we can see our GRAD algorithm have good performance in recognizing most activities. However, the Miss Rate for some activities such as *Fighting* and *Chase* are still high. This is because (a) the input features are very simple which are all derived from the MBB information, (b) the number of training samples for these activities is small, and (c) it is more difficult to correctly cluster the symmetric activities such as *Fighting* due to their large variance. Therefore, in order to further improve the performance, more sophisticated input features [19, 20] can be used and the methods to train models in case of insufficient training data can be introduced [9, 25]. Furthermore, Fig. 8 also shows a large FA rate in the activity *Ignore*. This is because *Ignore* is a generalized activity in our experiment. Since we model *Ignore* as the non-interaction case between people, it can be confused with all the other activities including both symmetric and asymmetric ones. This leads to the large number of samples misclassified as *Ignore*.

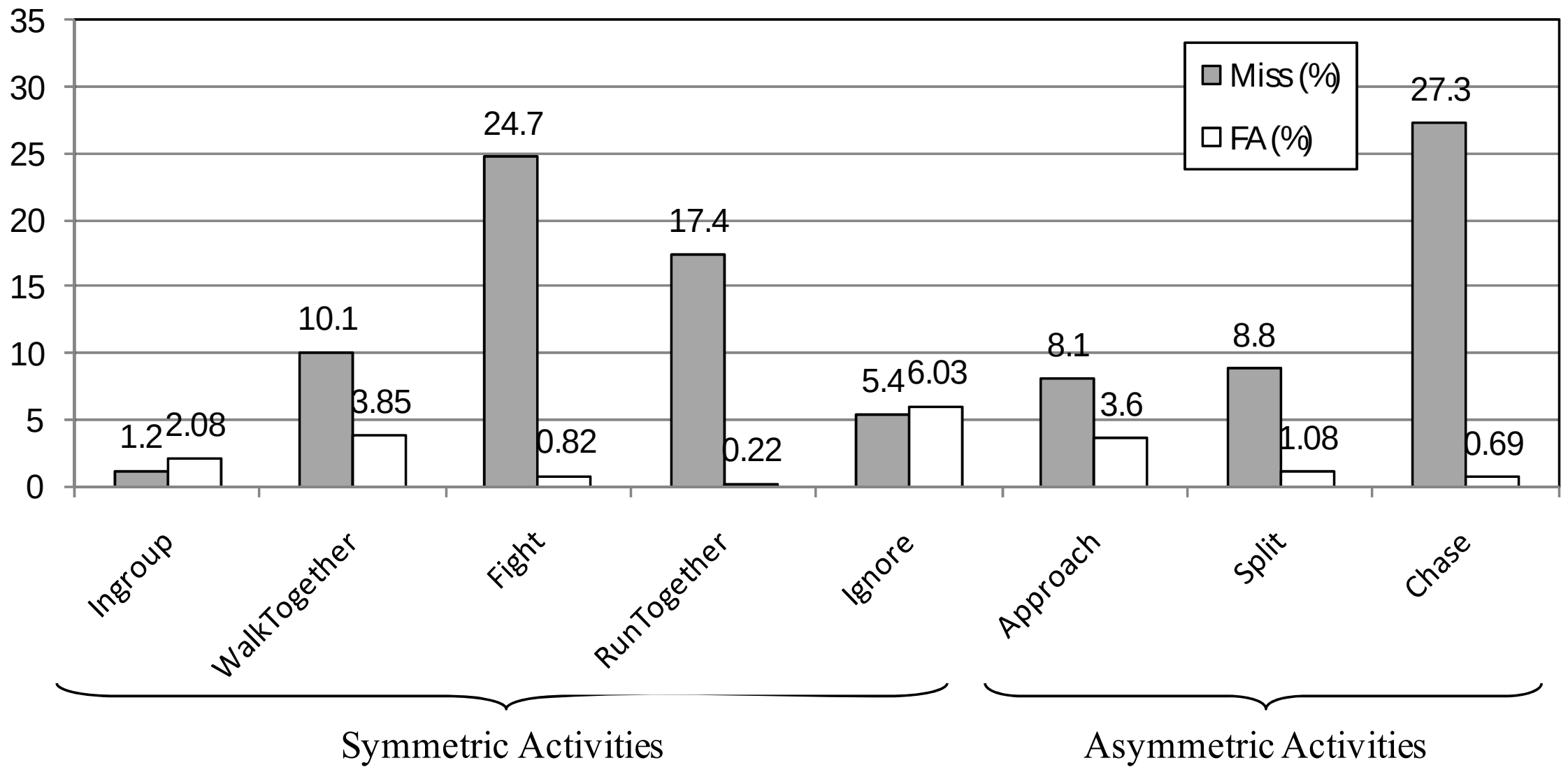

Fig. 8. The average Frame Level FA and Miss for the GRAD algorithm.

## VII. Algorithm Extension

In the GRAD algorithm proposed in this paper, we model hierarchical-structure activities based on our Symmetric-Asymmetric Activity Structure (SAAS) and cluster people into symmetric sub-groups based on the SRC clustering algorithm. The higher level asymmetric activities between symmetric sub-groups can then be recognized based on the interactions between Group Representatives for each symmetric sub-group. We believe that the framework of our proposed GRAD algorithm is general and can easily be extended. In this section, we discuss some possible extensions of our GRAD algorithm.

1) In this paper, we use *SAAS* to model hierarchical activities as a two-level structure with symmetric activities as the lower level and asymmetric activities as the higher level. This two-level structure can cover many scenarios in daily life. However, as mentioned, there may be activities with other hierarchical structures. For example, one *approaching* group may *chase* another *splitting* group and these two asymmetric groups will form a *super asymmetric group*. In these cases, we can extend our Group Representative method so that GRs can also be calculated and used to represent asymmetric groups. Furthermore, we can also extend our SAAS to model different activity structures. In the above example, we can first extend our SAAS by adding one more asymmetric activity level over the original asymmetric level to form a Symmetric-Asymmetric-Asymmetric Activity Structure. The *chase* activity can then be recognized based on the interactions between the two GRs of the two asymmetric sub-groups of *approaching* and *splitting*.
2) In the experiments of this paper, all asymmetric activities take place only between two symmetric sub-groups. However, there may be cases that the asymmetric activities take place among three or more entities. For example, person *A* is *approaching* the symmetric sub-group *B*, at the same time, another person *C* is also *approaching* group *B* from another direction. These three symmetric subgroups *A*, *B* and *C* will form an asymmetric group of *approaching*. In these cases, we can extend our SRC clustering method to further cluster symmetric subgroups into asymmetric groups. In the above example, we can first calculate the *distance metrics* between *A*, *B* and *C* based on their asymmetric interaction, and then cluster them into one asymmetric group.

3) In this paper, we use AHMM to model the activity correlation metric between any two people, use our SRC clustering method to cluster people into symmetric sub-groups, and use one of the three proposed GRs (P-GR, V-GR and SV-GR) to represent each symmetric sub-group. However, since the framework of our GRAD algorithm is general, other models, clustering methods, and GR calculation methods can also be used to improve the performance of the GRAD method.

## VIII. Conclusion

In this paper, we proposed (a) a Symmetric-Asymmetric Activity Structure for the detection of hierarchical activities, (b) a Group Representative to handle the group event detection with a varying number of group members, and (c) an SRC clustering algorithm to deal with clustering with an asymmetric distance metric. Experimental results demonstrate the effectiveness of our proposed algorithm.

## Acknowledgment

This work was supported in part by the following grants: ARO PECASE Grant (W911NF-05-1-0491) and ARO MURI Grant (W 911 NF 0710287). The authors would like to thank Dr. Samy Dengio for providing part of the code for implementing the AHMM.

## References

[1] D. Zhang, D. Gatica-Perez, S. Bengio and I. McCowan, "Modeling individual and group actions in meetings with layered HMMs," *IEEE Trans. Multimedia*, vol. 8, pp. 509-520, 2006.

[2] N. Oliver, E. Horvitz, and A. Garg. "Layered representations for learning and inferring office activity from multiple sensory channels," *Proc. ICMI*, October 2002.

[3] S. Park and J.K. Aggarwal, "A hierarchical bayesian network for event recognition of human actions and interactions," *Association for Computing Machinery Multimedia Systems Journal*, 2004.

[4] S. Hongeng and R. Nevatia, "Multi-agent event recognition," in *Proc. IEEE Int'l. Conf. Computer Vision*, July 2001.

[5] N. Vaswani, A.R. Chowdhury and R. Chellappa, "Activity recognition using the dynamic of the configurations of interacting objects," *IEEE Conf. Computer Vision and Pattern Recognition*, 2003.

[6] D. Wyatt, T. Choudhury and J. Bilmes, "Conversation detection and speaker segmentation in privacy-sensitive situated speech data," *Speech and audio processing for intelligent environments*, 2007.

[7] S. Bengio, "An asynchronous hidden Markov model for audio-visual speech recognition," S. *Proc. NIPS 15*, 2003.

[8] BEHAVE data, http://groups.inf.ed.ac.uk/vision/behavedata/interactions/.

[9] W. Lin, M.-T. Sun, R. Poovendran and Z. Zhang, "Activity Recognition using a Combination of Category Components and Local Models for Video Surveillance," *IEEE Trans. CSVT*, no. 8, 2008.

[10] H. Späth, "Cluster Analysis Algorithms for Data Reduction and Classification of Objects," *Halsted Press*, 1980.

[11] K. Smith, D. Gatica-Perez and J.M. Odobez, "Using Particles to Track Varying Number of Interacting People," *CVPR*, 2005.

[12] L. R. Rabiner and B.-H. Juang, "Fundamentals of Speech Recognition," *Prentice-Hall*, 1993.

[13] T. V. Duong, H. H. Bui, D. Q. Phung, and S. Venkatesh, "Activity recognition and abnormality detection with the switching hidden semi-Markov model," *Proc. IEEE Conf. Comput. Vis. Pattern Recognit.*, vol. 1, pp. 838–845, 2005.

[14] B. Li, E. Chang, and C. T. Wu, "DPF-a perceptual distance function for image retrieval," *Proc. Int. Conf. Image Process.*, vol. 2, pp. 597–600, 2002.

[15] F. Lv, J. Kang, R. Nevatia, I. Cohen, and G. Medioni, "Automatic tracking and labeling of human activities in a video sequence," *Proc. IEEE Workshop Performance Eval. Tracking and Surveillance*, pp. 33–40, 2004.

[16] I. McCowan, D. Gatica-Perez, S. Bengio, G. Lathoud, M. Barnard, and D. Zhang, "Automatic analysis of multimodal group actions in meetings," *IEEE Transactions on Pattern Analysis and Machine Intelligence*, vol. 27(3), pp. 305-317, 2005.

[17] L. Lam and S.Y. Suen, "Application of majority voting to pattern recognition: an analysis of its behavior and performance," *IEEE Transactions on Systems, Man and Cybernetics*, vol 27, issue 5, pp. 553 – 568, 1997.

[18] S.B. Oh, "On the relationship between majority vote accuracy and dependency in multiple classifier systems," *Pattern Recognitoin Letters*, vol. 24, pp. 359-363, 2003.

[19] Y. Song, L. Goncalves, and P. Perona, "Unsupervised learning of human motion," *IEEE Trans. Pattern Anal. Mach. Intell.*, vol. 25, no. 7, pp. 814–827, Jul. 2003.

[20] Y. A. Ivanov and A. F. Bobick, "Recognition of visual activities and interactions by stochastic parsing," *IEEE Trans. Pattern Anal. Mach. Intell.*, vol. 22, no. 8, pp. 852–872, Aug. 2000.

[21] A. Amer, "Voting-based simultaneous tracking of multiple video objects," *IEEE Trans. Circuits Syst. Video Technol.*, vol. 15, no. 11, pp.1448–1462, Nov. 2005.

[22] J. Bilmes, "A Gentle Tutorial of the EM Algorithm and Its Application to Parameter Estimation for Gaussian Mixture and Hidden Markov Models," *U.C. Berkeley*, *ICSI-TR-97-021*, 1997.

[23] P. C. Ribeiro and J. Santos-Victor, "Human activity recognition from video: Modeling, feature selection and classification architecture," in *Proc. Int. Workshop Human Activity Recognition and Modeling*, 2005, pp. 61–70.

[24] B. Moghaddam and A. Pentland, "Probabilistic visual learning for object representation," *IEEE Trans. Pattern Anal. Mach. Intel.*, vol. 19, no. 7, pp. 696–710, Jul. 1997.

[25] W. Lin, M.-T. Sun, R. Poovendran and Z. Zhang, "Human Activity Recognition for Video Surveillance," *ISCAS*, 2008.